\documentclass[10pt,twocolumn,letterpaper]{article}

\usepackage[review]{}      

\usepackage{graphicx}
\usepackage{amsmath}
\usepackage{amssymb}
\usepackage{booktabs}

\usepackage{times}
\usepackage{epsfig}
\usepackage{float}
\usepackage{xfrac}
\usepackage{makecell}
\usepackage{multicol}
\usepackage{multirow}
\usepackage{verbatim}
\usepackage{ulem}
\usepackage{bm}

\usepackage[pagebackref,breaklinks,colorlinks]{hyperref}

\usepackage[capitalize]{cleveref}
\crefname{section}{Sec.}{Secs.}
\Crefname{section}{Section}{Sections}
\Crefname{table}{Table}{Tables}
\crefname{table}{Tab.}{Tabs.}

\def\cvprPaperID{*****} 
\def\confName{}
\def\confYear{}

\begin{document}

\title{Valid Information Guidance Network for Compressed Video Quality Enhancement}

\author{Xuan Sun, Ziyue Zhang, Guannan Chen and Dan Zhu\\
BOE Technology Group Co., LTD.\\
{sunxuan@boe.com.cn}
}
\maketitle

\begin{abstract}

In recent years deep learning methods have shown great superiority in compressed video quality enhancement tasks. Existing methods generally take the raw video as the ground truth and extract practical information from consecutive frames containing various artifacts. However, they do not fully exploit the valid information of compressed and raw videos to guide the quality enhancement for compressed videos. 
In this paper, we propose a unique \textbf{V}alid \textbf{I}nformation \textbf{G}uidance scheme (VIG) to enhance the quality of compressed videos by mining valid information from both compressed videos and raw videos.
Specifically, we propose an efficient framework, Compressed Redundancy Filtering (CRF) network, to balance speed and enhancement. After removing the redundancy by filtering the information, CRF can use the valid information of the compressed video to reconstruct the texture.
Furthermore, we propose a progressive Truth Guidance Distillation (TGD) strategy, which does not need to design additional teacher models and distillation loss functions. By only using the ground truth as input to guide the model to aggregate the correct spatio-temporal correspondence across the raw frames, TGD can significantly improve the enhancement effect without increasing the extra training cost.
Extensive experiments show that our method achieves the state-of-the-art performance of compressed video quality enhancement in terms of accuracy and efﬁciency.

\end{abstract}

\section{Introduction}

\begin{figure}[t]
\centering
\includegraphics[width=1\linewidth]{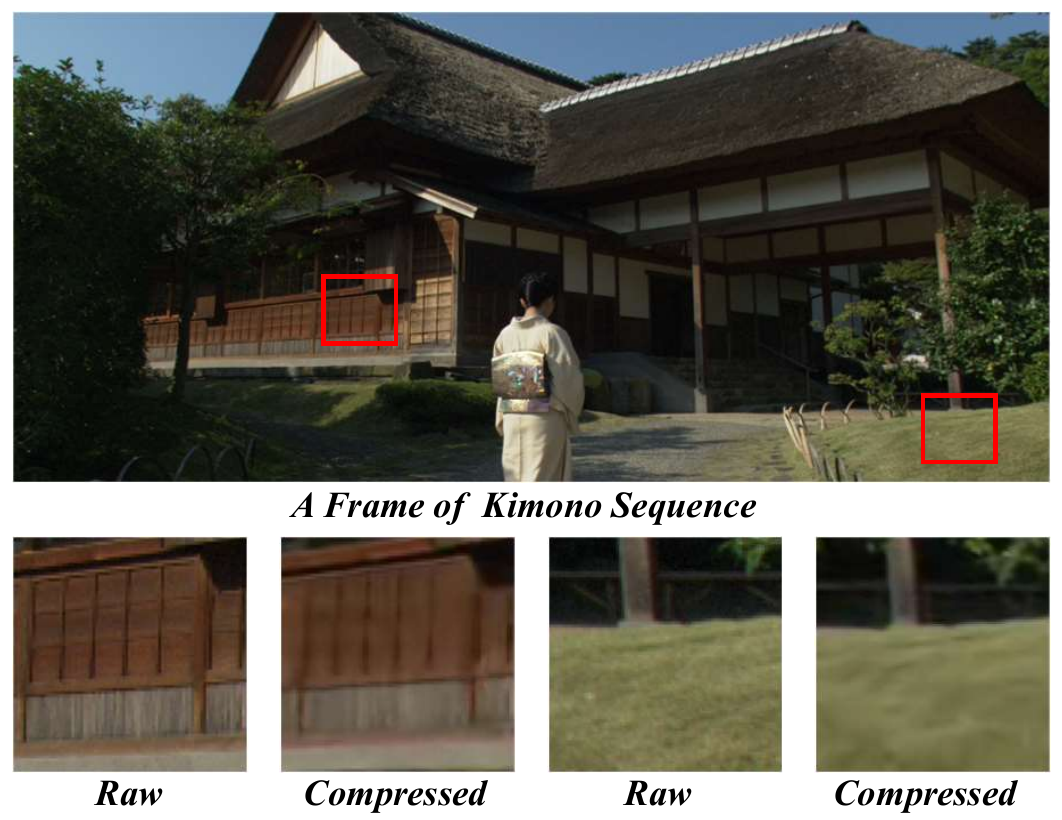}
\caption{\textbf{The valid information of the compressed patches and the raw patches.}
}
\label{fig:rongyu}
\end{figure}

Video data typically undergo a video encoding process for efficient transmission. However, existing block-based coding frameworks \cite{hevc,h264} often use inaccurate quantization and motion compensation techniques that lose high-frequency information and produce many compression artifacts, as shown in Figure \ref{fig:rongyu}. The subjective effect of compressed videos will be notably reduced, especially at a low bit-rate. 
We calculate the mean value of the pixel variance of each 64×64 patch of 18 standard test sequences\cite{jctvc} in Table \ref{tab:valid}. Significantly, the mean-variance value of the raw video is larger than that of the compressed video under the Low-Delay (LD) configuration. In other words, the compressed frames contain less valid information than the raw frames.
Moreover, low-quality compressed videos will severely impact the downstream tasks (e.g., classification, detection, segmentation).
With the mighty computing power of the terminal, the quality of the video passing through the transmission can be enhanced at the decoding end without increasing the amount of transmitted data. 
This dramatically boosts the subjective quality of the decoded video under the limited transmission bandwidth resources. Accordingly, it is critical to study compressed video quality enhancement (VQE) tasks.

Most current works \cite{mfqe1,mfqe2,stdc,RFDA,arcnn,dncnn,livqe,dcad,dscnn} focus on designing huge models structure to improve the quality of compressed videos. However, these works generally follow the design ideas of video enhancement tasks, such as video super-resolution reconstruction, which are not fully designed according to the characteristics of compressed videos. Some works\cite{mfqe1,mfqe2,stdc} introduce cost-efficient network architectures to reduce the computational burden and required memory.
Although these works achieve the balance of speed and effect, specially designed or complicated architectures may pose challenges to implementation on hardware devices.

Knowledge distillation is a model compression method that can assist the training process of a student network with the complementary knowledge. It can significantly improve the performance of the model without changing the structure.
PISR\cite{PISR} is a known knowledge distillation method for image super-resolution reconstruction. It inherits the general knowledge distillation method and needs to design a vast teacher model to assist the student model, which significantly increases the training cost. To our best knowledge, there is no knowledge distillation method for compressed video enhancement. 

In this paper, we propose a novel Valid Information Guidance (VIG) scheme to accomplish the compressed video enhancement task. The main idea of VIG is to fully exploit the valid cues in compressed and original videos to guide video reconstruction. In particular, we first develop an effective Compressed Redundancy Filtering (CRF) network to capture the most valid context by excluding redundant content. In addition, we devise a counter-intuitive Truth Guidance Distillation (TGD) strategy to enhance the reconstruction progressively by understanding the original pixel distribution in the original video. In summary, our contributions are as follows:

\begin{itemize}
\item We propose a novel Valid Information Guidance (VIG) scheme to model the spatio-temporal dependency for compressed video enhancement tasks. 

\item Comprehensive experiments are exhibited to explore the effect of the valid information flow between raw videos and compressed videos. 

\item VIG outperforms contemporary methods and demonstrates the new state-of-the-art performance on the VQE benchmark dataset.
\end{itemize}

\section{Related Work}
\paragraph{Redundancy Filtering Method}
Over the past decade, many studies \cite{stdc,mfqe1,mfqe2,RFDA,ali-icassp} have proved that using inter-frame information has a crucial impact on the results of VQE.
From the perspective of information extraction, VQE techniques can be divided into two categories: the direct extraction method and the redundancy filtering method.
The direct extraction method refers to the operation of direct aggregating information in the original input, such as STDF\cite{stdc}, MFQE2\cite{mfqe2} and RFDA\cite{RFDA}. Correspondingly, the redundancy filtering method refers to the operation of compressing information before the primary aggregation. 
Down-sampling is common in classification, segmentation, and other high-level tasks in computer vision \cite{unet,unet++,vit,swin} to generate high-level representations.
Nevertheless, it is rarely used in image or video enhancement tasks because it may result in the loss of many details and affect the reconstruction quality. 
HUPN\cite{HPUN} proposes a framework that introduces the down-sampling operation into image super-resolution, representing the possibility of compressing information before the primary aggregation in the image enhancement task. 
STDF\cite{stdc} employs a module like Unet\cite{unet} to predict the offset between compressed frames and does not consider the up-down sampling structure as the overall framework.  
Noted that compared with the image to be super-resolution, the compressed video has more redundant information to be filtered. In this work, we propose CRF to exploit spatio-temporal information more effectively and efficiently via excluding the redundant content several times. Note that compared with super-resolution images, the compressed video has more redundant information to filter. In this work, we propose the CRF module to effectively exploit spatio-temporal information by eliminating redundant content several times.

\begin{figure*}[t]
\centering
\includegraphics[width=1\linewidth]{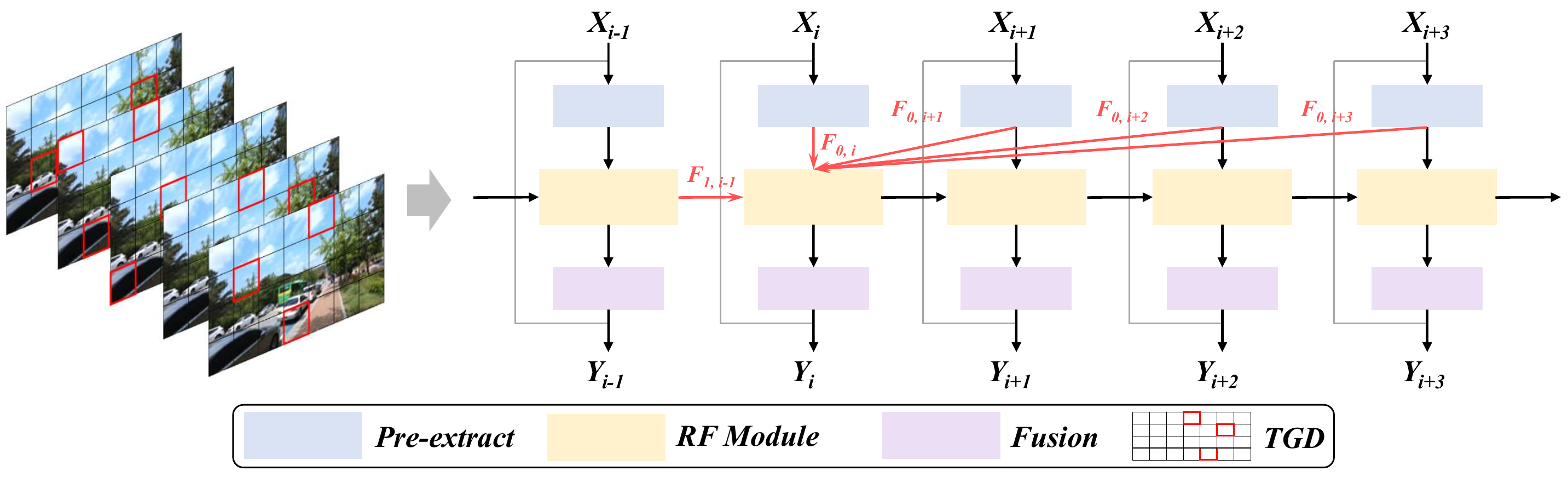} 
\caption{\textbf{An overview of VIG.} VIG generally follows a single propagation approach. Specifically, it consists of two parts: Truth Guidance Distillation (TGD) strategy and Compressed Redundancy Filtering (CRF) network. 
For TGD, the red box exploits the corresponding compressed image region, and the black box means using the corresponding areas of GT. For CRF, Redundancy Filtering (RF) module is the key part to capture the valid spacia-temporal information. For TGD, the red box exploits the corresponding compressed image region, and the black box uses the corresponding areas of GT. For CRF, the Redundancy Filtering (RF) module is crucial to capture valid spatio-temporal information.
}
\label{fig:VIG}
\end{figure*}

\begin{table}[t]
\centering
{
\resizebox{0.95\columnwidth}{!}
{%
\begin{tabular}{c|c c c|c}
\hline
\multicolumn{1}{c|}{Mode} &
\multicolumn{1}{c}{ } &
\multicolumn{1}{c}{LD} &
\multicolumn{1}{c|}{ } &
\multicolumn{1}{c}{RAW} \\
\hline
QP & 32 & 37 & 42 & -  \\
\hline
Mean Value & 1099 & 1075 & 1032 & 1230 \\
\hline
\end{tabular}}
}
\caption{\textbf{The mean value of the pixel variance of 18 standard test sequences\cite{jctvc}.} The smaller mean-variance indicates the flatter content and less textures retained within pixel patches.}
\label{tab:valid}
\end{table}

\paragraph{Valid Information Application Strategy}
Generative self-supervised learning for computer vision has achieved tremendous progress. Several works\cite{MAE,simmim,videomae,A2MIM,mixmim} focus on using valid information via the pre-text task to improve downstream vision tasks. Several works\cite{MAE,simmim,videomae,A2MIM,mixmim} focus on improving downstream visual tasks by using effective information in pre-text tasks.
In detial, MAE\cite{MAE} and SimMIM\cite{simmim} replace a random subset of input tokens with a special MASK symbol and aim at reconstructing original image tokens from the corrupted image with Vision transformers\cite{vit,swin}. In detail, MAE\cite{MAE} and SimMIM\cite{simmim} replace a random subset of input tokens with a special MASK symbol designed to reconstruct the original image tokens from corrupted images using Vision transformers\cite{vit,swin}. Subsequently, VideoMAE\cite{videomae} proves that an extremely high proportion of masking ratio still yields favorable performance on videos. A2MIM\cite{A2MIM} adopts a masked image modelling method on convolutional neural networks (CNNs). Further, MixMIM\cite{mixmim} finds that using the mask symbol significantly causes training-finetuning inconsistency and replaces the masked tokens of one image with visible tokens of another image. 
Notably, all the above methods are unsuitable for VQE because there is an apparent gap between reconstructing the masked region and enhancing the compressed region\cite{cutblur}. 
CutBlur\cite{cutblur} proposes a data augmentation method that cuts a low-resolution patch and pastes it to the corresponding high-resolution image region. Nevertheless, it only cuts and pastes randomly in the image task and does not fully exploit the guidance of Ground Truth in the video task.
PISR\cite{PISR} designs a distillation framework by using ground-truth high-resolution images as privileged information, which is most similar to our work. 
However, it follows the traditional distillation method and designs an additional teacher network with an imitation loss, significantly increasing the training cost. Inspired by PISR\cite{PISR}, we propose a counter-intuitive knowledge distillation strategy by exploiting the raw frames as the input with little increase in computing costs.

\section{Methodology}

\subsection{Overall Architecture}
\label{section:overview}

We train the model in two stages: pre-training and fine-tuning. 
Moreover, the overall architecture of VIG can be divided into two parts: Truth Guidance Distillation (TGD) strategy and Compressed Redundancy Filtering (CRF) network. 
We assume that $\mathbf{X}_{i}$ is the input frame and $\mathbf{Y}_{i}$ is the output frame at time $i$. A total of $2k+1$ frames will be used to compute a video clip.

These input frames are denoted by 
$\mathbf{X}_{t}=\{ \mathbf{X}_{i-k}, \mathbf{X}_{i-k+1}, \ldots, \mathbf{X}_{i}, \ldots, \mathbf{X}_{i+k-1}, \mathbf{X}_{i+k}\}$. 
We represent the enhanced frames by $\mathbf{Y}_{t}=\{ \mathbf{Y}_{i-k}, \mathbf{Y}_{i-k+1}, \ldots, \mathbf{Y}_{i}, \ldots, \mathbf{Y}_{i+k-1}, \mathbf{Y}_{i+k}\}$. The enhanced frame $\mathbf{Y}_{t}$ can be generated by
    \begin{align}
        \label{equ:intra_BB_propagation}
        \mathbf{Y}_{t}  = \mathit{V}_{\theta}\left(\mathbf{X}_{t} \right),
    \end{align}
where $\mathit{V}_{\theta}(\cdot)$ represents the whole process of VIG. $\mathit{\theta}$ is the learnable parameters of VIG. Specifically, taking $\mathbf{Y}_{i}$ as an example, the enhancement process can be formalized as
    \begin{align}
        \label{equ:intra_BB_propagation}
        \mathbf{Y}_{i},\mathbf{F}_{1,i}  = \mathit{V}_{\theta}\{\mathbf{X}_{i},\mathbf{X}_{i+1},\mathbf{X}_{i+2},\mathbf{X}_{i+3},\mathbf{F}_{1,i-1} \},
    \end{align}
where $\mathbf{F}_{1,i-1}$ represents the hidden features of all enhanced frames generated by the RF module. 

\subsection{TGD strategy}

\begin{figure}[ht]
\centering
\includegraphics[width=1\linewidth]{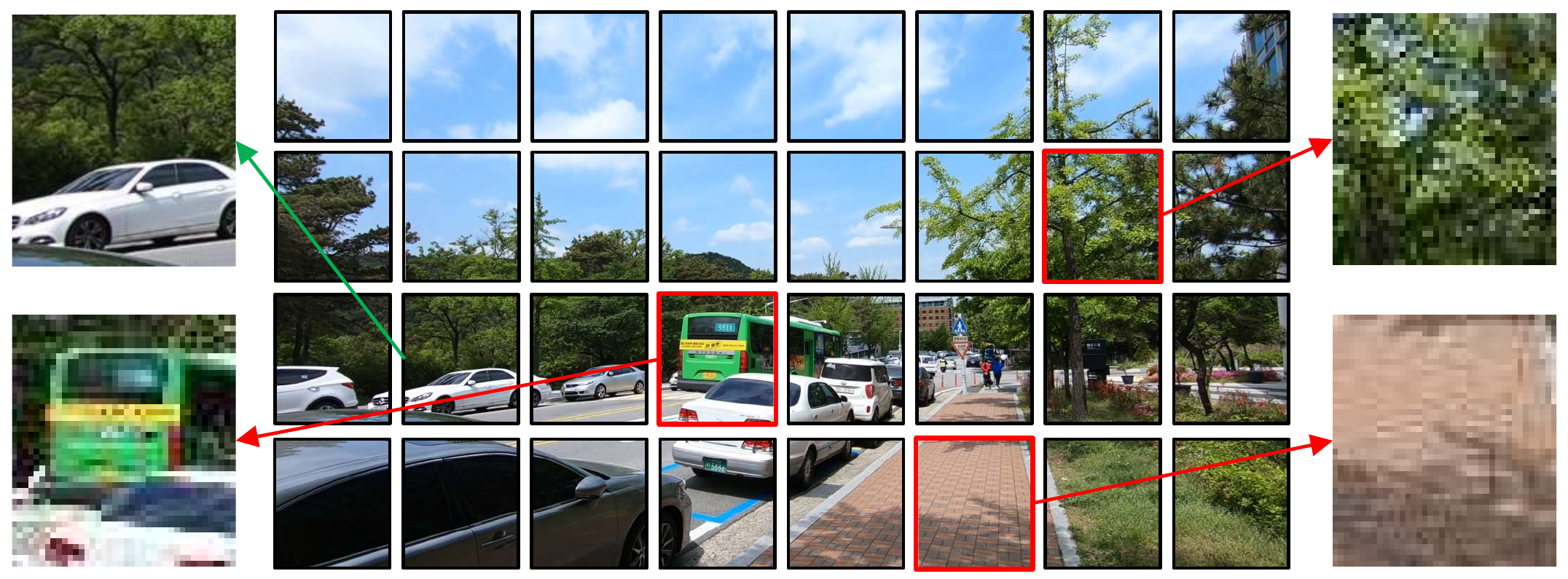}
\caption{\textbf{TGD strategy}. Three red boxes are the compressed patches added into GT; the other boxes represent the raw patches.
}
\label{fig:TGD}
\end{figure}

The goal of TGD is to generate new training samples by cut-and-pasting the random region of $\mathit{f}_{GT}$ into the corresponding $\mathit{f}_{C}$. We adopt a patch-based random masking strategy. The input of the pre-training stage is defined as  
\begin{align}
        \label{equ:mask}
        \mathbf{X}_{i} = \mathit{M}_{r,w}\mathit{\odot}\mathit{f}_{i}^{GT}+\left(1-\mathit{M}_{r,w}\right)\mathit{\odot}\mathit{f}_{i}^{C},
    \end{align}
where $\mathit{M}_{r,w}$ is the selected region of the corresponding $\mathit{f}_{i}^{GT}$. $\mathit{r}$ is the ratio of $\mathit{f}_{i}^{GT}$ in $\mathbf{X}_{i}$, $\mathit{w}$ is the size of the mask in $\mathbf{X}_{i}$, and $\mathit{\odot}$ is element-wise multiplication. Figure \ref{fig:TGD} shows the progress of TGD. We cut $\mathit{f}_{i}^{GT}$ into $\mathit{n}$ patches according to $\mathit{w}$, and randomly replace several patches with the corresponding regions of $\mathit{f}_{i}^{C}$ on the basis of $\mathit{r}$. We consider patch sizes of different resolution stages, from 2×2 to 32×32.

Since the substituted token is chosen randomly in $\mathit{f}_{i}^{GT}$, TGD can encourage a model to enjoy the regularization effect by learning the local and global relationships among pixels. $\mathit{r}$ and $\mathit{w}$ can be changed as needed. $\mathit{r}$ used in this paper is set to 0.9. In other words, the TGD strategy almost entirely uses GT as input. Due to the computational requirements of back-propagation, we only add a tiny ratio of $\mathit{f}_{i}^{C}$ into GT. 
Moreover, we set $\mathit{w}$ to 32, 16, 8, and 4 for four periods to progressively learn the pixel relationship from global to local. We train VIG for 5 thousand iterations in every period.

Significantly, we only use TGD in the pre-training phase. So the input of the fine-tuning stage can be considered as
\begin{align}
        \label{equ:intra_BB_propagation}
        \mathbf{X}_{i} = \mathit{f}_{i}^{C}.
    \end{align}

\subsection{CRF Network}
As shown in Figure \ref{fig:VIG}, CRF contains a recurrent forward network. It consists of three parts: A Pre-extract module, a Redundancy Filtering (RF) module and a Fusion module. 
The Pre-extract module is targeted for gaining the primitive features of input frame $\mathbf{X}_{i}$.
For primitive features, at the $i$-th timestamp, let $\mathit{P}_{\alpha}(\cdot)$ denote the feature extract module, $\mathit{R}_{\beta}(\cdot)$ denote the RF module and $\mathbf{F}_{0,i}$ represent the output of $\mathit{P}_{\alpha}(\cdot)$. 
More specifically, the RF module receives the primitive features from current and future frames $\{\mathbf{X}_{i}, \ldots, \mathbf{X}_{i+k-1}, \mathbf{X}_{i+k}\}$ and gets the hidden features to exploits all the information of the past frames $\{\mathbf{X}_{i-k}, \mathbf{X}_{i-k+1}, \ldots, \mathbf{X}_{i-1}\}$. Then we take advantage of $\mathit{F}_{\gamma}(\cdot)$ to be the Fusion module to generate the output $\mathbf{Y}_{i}$. The operation of CRF can be formalized as 
\begin{align}
        \label{equ:pre-extrac}
        \mathbf{F}_{0,i}=\mathit{P}_{\alpha}\left (\mathbf{X}_{i}\right ),
    \end{align}
    \begin{align}
        \label{equ:RF}
        \mathbf{F}_{1,i}  = \mathit{R}_{ \beta}\{\mathbf{F}_{0,i},\mathbf{F}_{0,i+1},\mathbf{F}_{0,i+2},\mathbf{F}_{0,i+3},\mathbf{F}_{1,i-1} \},
    \end{align}
\begin{align}
        \label{equ:fusion}
        \mathbf{Y}_{i}=\mathit{F_{\gamma}}\left (\mathbf{F}_{1,i}\right )+\mathbf{X}_{i},
    \end{align}
where $\mathit{\alpha}$, $\mathit{\beta}$ and $\mathit{\gamma}$ are the learnable parameters. To keep our design paradigm concise, $\mathit{P}_{\alpha}(\cdot)$ and $\mathit{F}_{\gamma}(\cdot)$ are each implemented with a single convolution layer.

\begin{figure}[tb]
\centering
\includegraphics[width=1\linewidth]{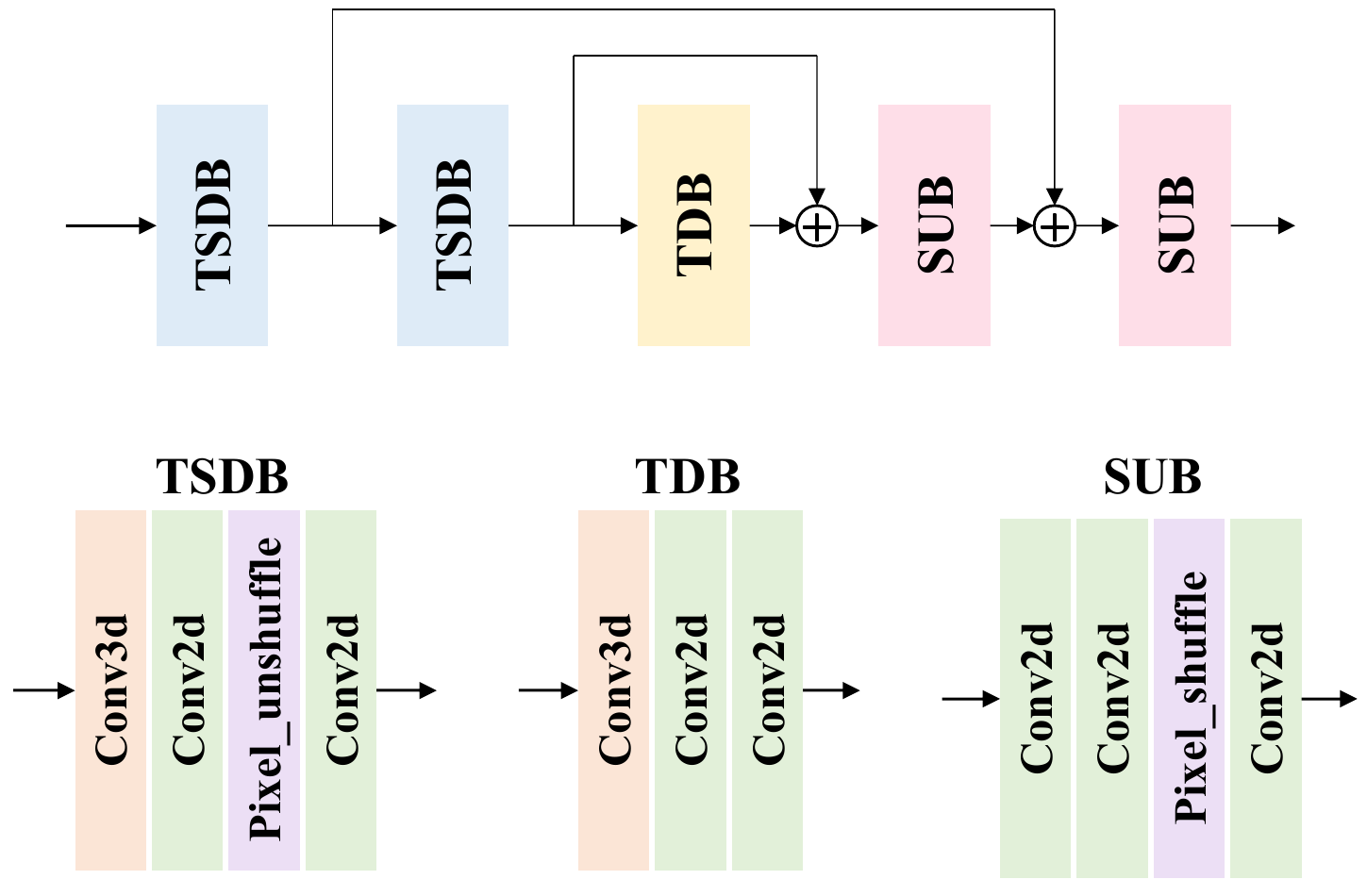}
\caption{\textbf{RF module}. RF consists of three parts: the Temporal-spacial Downsampling  (TSD) block, the Temporal Downsampling (TD) block, and the Spacial Upsampling (SU) block. 
}
\label{fig:RF}
\end{figure}

\textbf{RF module.}
AS shown in Figure \ref{fig:RF}, We employ the TSD block $\mathit{D}_{\delta}^{TSD}(\cdot)$ and the TD block $\mathit{D}_{\eta}^{TD}(\cdot)$ to filter the spatio-temporal redundancy, and then utilize the SU block $\mathit{U}_{\mu}^{SU}(\cdot)$ to recover the resolution. 
The downsampling operation of the RF module can be formalized as
\begin{align}
        \label{equ:RF}
        \mathbf{F}_{TSD1,i}  = \mathit{D}_{\delta1}^{TSD}\{\mathbf{F}_{0,i},\mathbf{F}_{0,i+1},\mathbf{F}_{0,i+2},\mathbf{F}_{0,i+3},\mathbf{F}_{1,i-1} \},
    \end{align}
\begin{align}
        \label{equ:RF}
        \mathbf{F}_{TSD2,i}  = \mathit{D}_{\delta2}^{TSD}\left (\mathbf{F}_{TSD1,i}\right ),
    \end{align}
\begin{align}
        \label{equ:RF}
        \mathbf{F}_{TD,i}  =\mathit{D}_{\eta}^{TD}\left (\mathbf{F}_{TSD2,i}\right ),
    \end{align}
where $\mathit{\delta1}$, $\mathit{\delta2}$, and $\mathit{\eta}$ are the learnable parameters.
The upsampling operation of the RF module can be described as 
\begin{align}
        \label{equ:RF}
        \mathbf{F}_{SU1,i}  =\mathit{U}_{\mu1}^{SU}\left( \mathbf{F}_{TD,i}+\mathbf{F}_{TSD2,i} \right ),
    \end{align}
\begin{align}
        \label{equ:RF}
        \mathbf{F}_{SU2,i}  =\mathit{U}_{\mu2}^{SU}\left( \mathbf{F}_{TD,i}+\mathbf{F}_{TSD2,i} \right ),
    \end{align}
where $\mathit{\mu1}$ and $\mathit{\mu2}$ are the learnable parameters.

\subsection{Training Scheme}
In both stages, we employ the Charbonnier Loss $\pounds$ \cite{loss} to train the model. 
In the first stage, using TGD, only the Charbonnier loss between the compressed patches and the corresponding raw patches is calculated. This way, VIG can better learn the flow of valid information in GT. The pre-training process can be formalized as
\begin{align}
        \label{equ:pretraining}
        \pounds = \sqrt{(\left(1-\mathit{M}_{r,w}\right)\mathbf{Y}_{i}-\left(1-\mathit{M}_{r,w}\right)\mathit{f}_{i}^{GT})^{2}+\epsilon},
    \end{align}
where $\epsilon$ is set to ${10}^{-6}$ here.
In the second stage, following previous works, we use $\pounds$ between $\mathbf{Y}_{i}$ and $\mathit{f}_{i}^{GT}$ to optimize the model. 
\begin{align}
        \label{equ:fine-tuning}
        \pounds = \sqrt{(\mathbf{Y}_{i}-\mathit{f}_{i}^{GT})^{2}+\epsilon}.
    \end{align}
Note that only 20 thousand iterations are needed for the pre-training stage, and the fine-tuning stage costs another 100 thousand iterations in this paper. 

\section{Experiments}
\label{section:experiments}

\begin{table*}[ht]
\centering
{
\resizebox{0.95\textwidth}{!}
{%
\begin{tabular}{c|c|c|c|c|c|c|c|c}
\hline
\multicolumn{1}{c|}{Class}&
\multicolumn{1}{c|}{Sequence} &
\multicolumn{1}{c|}{AR-CNN} &
\multicolumn{1}{c|}{DnCNN} &  
\multicolumn{1}{c|}{RNAN} & 
\multicolumn{1}{c|}{MFQE2.0} &
\multicolumn{1}{c|}{STDF-R3} &
\multicolumn{1}{c}{Ours}\\
\hline
\multirow{2}*{\makecell[c]{A\\(2560*1600)}}
& PeopleOnStreet& 0.37/0.76 & 0.54/0.94 & 0.74/1.30 & 0.92/1.57 &  1.18/1.82 & \textbf{1.30}/\textbf{2.13}  \\
&Traffic        & 0.27/0.50 & 0.35/0.64 & 0.40/0.86 & 0.59/1.02	& 0.65/1.04 & \textbf{0.74}/\textbf{1.24} \\
\hline
\multirow{5}*{\makecell[c]{B\\(1920*1080)}}
& Kimono & 0.20/0.59 & 0.27/0.73 & 0.33/0.98 & 0.55/1.18 &  0.77/1.47 & \textbf{0.92}/\textbf{1.74} \\
&ParkScene	& 0.14/0.44 & 0.17/0.52 & 0.20/0.77 & 0.46/1.23 &  0.54/1.32 & 0.55/1.44 \\
&Cactus	& 0.20/0.41 & 0.28/0.53 & 0.35/0.76 & 0.50/1.00 &  0.70/1.23 & \textbf{0.73}/\textbf{1.37} \\
&BQTerrace	& 0.23/0.43 & 0.33/0.53 & 0.42/0.84 & 0.40/0.67 &  0.58/0.93 & \textbf{0.55}/\textbf{0.99} \\
&BasketballDrive & 0.23/0.51 & 0.33/0.63 & 0.43/0.94 & 0.47/0.83 &  0.66/1.07 & \textbf{0.77}/\textbf{1.35}\\
\hline
\multirow{4}*{\makecell[c]{C\\(832*480)}}
&RaceHorses	&0.23/0.49 & 0.31/0.70 & 0.35/0.99 & 0.39/0.80 &  0.48/1.09 & 0.50/1.31\\
&BQMall&     0.28/0.69 & 0.38/0.87 & 0.45/1.15 & 0.62/1.20 &  0.90/1.61 & \textbf{0.94}/\textbf{1.88} \\
&PartyScene& 0.14/0.52 & 0.22/0.69 & 0.30/0.98 & 0.36/1.18 &  0.60/1.60 & \textbf{0.63}/\textbf{2.00}  \\
&BasketballDrill& 0.23/0.48 & 0.42/0.89 & 0.50/1.07 & 0.58/1.20 &  0.70/1.26 & \textbf{0.86}/\textbf{1.62} \\

\hline
\multirow{4}*{\makecell[c]{D\\(416*240)}}
&RaceHorses& 0.26/0.59 & 0.34/0.80 & 0.42/1.02 & 0.59/1.43 &  0.73/1.75 & 0.76/2.07 \\
&BQSquare&   0.21/0.30 & 0.30/0.46 & 0.32/0.63 & 0.34/0.65 &  0.91/1.13 & 0.86/\textbf{1.36} \\
&BlowingBubbles& 0.16/0.46 & 0.25/0.76 & 0.31/1.08 & 0.53/1.70 &  0.68/1.96 & \textbf{0.73}/\textbf{2.39} \\
&BasketballPass& 0.26/0.63 & 0.38/0.83 & 0.46/1.08 & 0.73/1.55 &  0.95/1.82 & \textbf{1.07}/\textbf{2.31} \\
\hline
\multirow{3}*{\makecell[c]{E\\(1280*720)}}
&FourPeople& 0.40/0.56 & 0.54/0.73 &0.70/0.97 & 0.73/0.95 &  0.92/1.07 & \textbf{1.06}/\textbf{1.33} \\
&Johnny& 0.24/0.21 & 0.47/0.54 & 0.56/0.88 & 0.60/0.68 &   0.69/0.73 & \textbf{0.90}/\textbf{1.09} \\ 
&KristenAndSara& 0.41/0.47 & 0.59/0.62 & 0.63/0.80 & 0.75/0.85 &  0.94/0.89 & \textbf{1.09}/\textbf{1.13} \\
\hline
\multirow{4}*{LD}
&QP37 Average& 0.25/0.50 & 0.36/0.69 & 0.44/0.95 & 0.56/1.09 &  0.75/1.32 & \textbf{0.83}/\textbf{1.60} \\
&QP32 Average& 0.19/0.17 & 0.33/0.41 & 0.41/0.62 & 0.52/0.68 &  0.73/0.87 & 0.77/1.03\\
&QP27 Average& 0.16/0.09 & 0.33/0.26 & - & 0.49/0.42 &  0.67/0.53 &0.73/0.64\\
&QP22 Average& 0.13/0.04 & 0.27/0.14 & - & 0.46/0.27 &  0.57/0.30 & 0.63/0.36\\
\hline
LD&Total& 0.18/0.20 & 0.32/0.38 &-& 0.51/0.62 & 0.68/0.76 & 0.74/0.91 \\

\hline
\end{tabular}

}
}
\vspace{3pt}
\caption{\textbf{Quantitative results of $\Delta$PSNR (dB) / $\Delta$SSIM ($\times{10}^{-2}$) on test videos at 4 different QPs.} The specific value of each sequence in the table is measured when the QP is 37. We show the average values of each method with 4 QPs.}
\label{tab:comparison of VIG model with others}
\end{table*}

To show the effectiveness and superiority of VIG, we have conducted meticulous experiments on the MFQE 2.0 dataset following by \cite{mfqe2,stdc,RFDA}. In addition, we perform extensive ablation studies to analyze the importance of each component of VIG and to understand it comprehensively.


\subsection{Datasets}
MFQE 2.0 dataset contains 126 videos with large range of resolutions: SIF (352×240), CIF (352×288), NTSC (720×486), 4CIF (704×576),240p (416×240), 360p (640×360), 480p (832×480), 720p (1280×720), 1080p (1920×1080), and WQXGA (2560×1600). 106 videos of them are selected for training and the rest are for validation. For testing, we adopt 18 standard test sequences of Joint Collaborative Team on Video Coding (JCT-VC) \cite{jctvc} as the test set.
This test dataset covers different scene conditions and can better verify the robustness of different approaches widely used in developing HEVC standards. All sequences are encoded in HEVC Low-Delay-P (LDP) configuration, using HM 16.20 with four different Quantization Parameters (QPs), i.e., 22, 27, 32 and 37 \cite{mfqe2}.

\subsection{Experimental Settings}
\label{paragraph:Experimental Settings}
VIG is implemented based on the PyTorch framework and trained with NVIDIA GeForce RTX 3090 GPUs. We use Adam optimizer \cite{adam} with $\beta_{1}=0.9$, $\beta_{2}=0.99$, and use Cosine Annealing \cite{sgdr} to decay the learning rate from $3\times{10}^{-4}$ to ${10}^{-4}$. 
We randomly crop image clips from the raw videos and the corresponding compressed videos as training samples for training. The image size is 128 × 128, and the batch size is 32. 
We also adopt flip and rotation as data augmentation strategies to further expand the dataset. We split each video in the training set into several video clips containing 13 frames. 
We only report quality enhancement on Y-channel in YUV/YCbCr space for evaluation.
We use Peak Signal-to-Noise Ratio (PSNR) and Structural Similarity Index (SSIM) \cite{ssim} to evaluate the quality of images generated by the VQE methods.

\subsection{Comparisons with State-of-the-art Methods}
\subsubsection{Quantitative Evaluation}
We compare VIG to the following state-of-the-art VQE models, like AR-CNN \cite{arcnn},  DnCNN\cite{dncnn},  RNAN\cite{rnan}, MFQE1.0 \cite{mfqe1},MFQE2.0\cite{mfqe2}, MRDN\cite{ali-icassp}, and STDF \cite{stdc}. 
The quantitative results are summarized in Table \ref{tab:comparison of VIG model with others} and the speed and performance comparison are also provided in Table \ref{tab:speed}. 
As shown in Table \ref{tab:comparison of VIG model with others}, VIG outperforms the existing state of the arts on MFQE2.0 datasets at four QPs, which proves the robustness of VIG.
Furthermore, VIG remarkably outperforms the other methods on most sequences, demonstrating its superior robustness and generalization capability. 
In particular, it gets +10.67$\%$/+0.08 dB higher $\Delta$PSNR than STDF-R3 \cite{stdc} and +6.41$\%$/+0.05 dB dB higher $\Delta$PSNR than MRDN \cite{ali-icassp} with QP 37.
It is important to note that our training volume is 1/3 of STDF-R3.

In addition, we use $\textit{FLOPs/frame}$ to measure computational complexity. The $\textit{FLOPs/frame}$ indicates FLOPs over per output frame.
VIG follows a bidirectional propagation method, so it can simultaneously generate multiple frames corresponding to the input. Therefore, the $\textit{FLOPs/frame}$ of VIG is 56.9 GB, showing great competitiveness. The parameter weight of VIG is 1.74 MB.

\begin{table}[tb]
\centering
{
\resizebox{0.95\columnwidth}{!}
{%
\begin{tabular}{c| c| c | c}
    \hline
    \multicolumn{1}{c|}{Method} &
    \multicolumn{1}{c|}{$\Delta$PSNR(dB)} &
    \multicolumn{1}{c|}{$\Delta$SSIM($\times{10}^{-2}$)} &
    \multicolumn{1}{c}{Speed (FPS)} \\
    \hline
    STDF-R3 & 0.75 & 1.32 & 59.0 \\
    \hline
    VIG & 0.83  & 1.60 & 60.8 \\
    \hline
    \end{tabular}
    }
}
\vspace{3pt}
\caption{\textbf{Speed of VIG.} 
Inference speed comparison between our method and STDF-R3. For a fair comparison, all methods are tested with 480p video, on a NVIDIA RTX 3090.The results are reported by frame per second (FPS). }
\label{tab:speed}
\end{table}

\textbf{Efficiency of VIG.} 
We compared VIG with STDF-R3\cite{stdc} as Table \ref{tab:speed}, who is the fastest VQE model.
Compared with STDF-R3, VIG runs +10.67$\%$/+0.08 dB with faster speed. 
Thanks to the design of UD, a great deal of computation is carried through on low-resolution feature maps. In addition, the output of VIG is continuous multiple frames, so the inference efficiency performs well.

\begin{figure}[tb]
\centering
\includegraphics[width=1\linewidth]{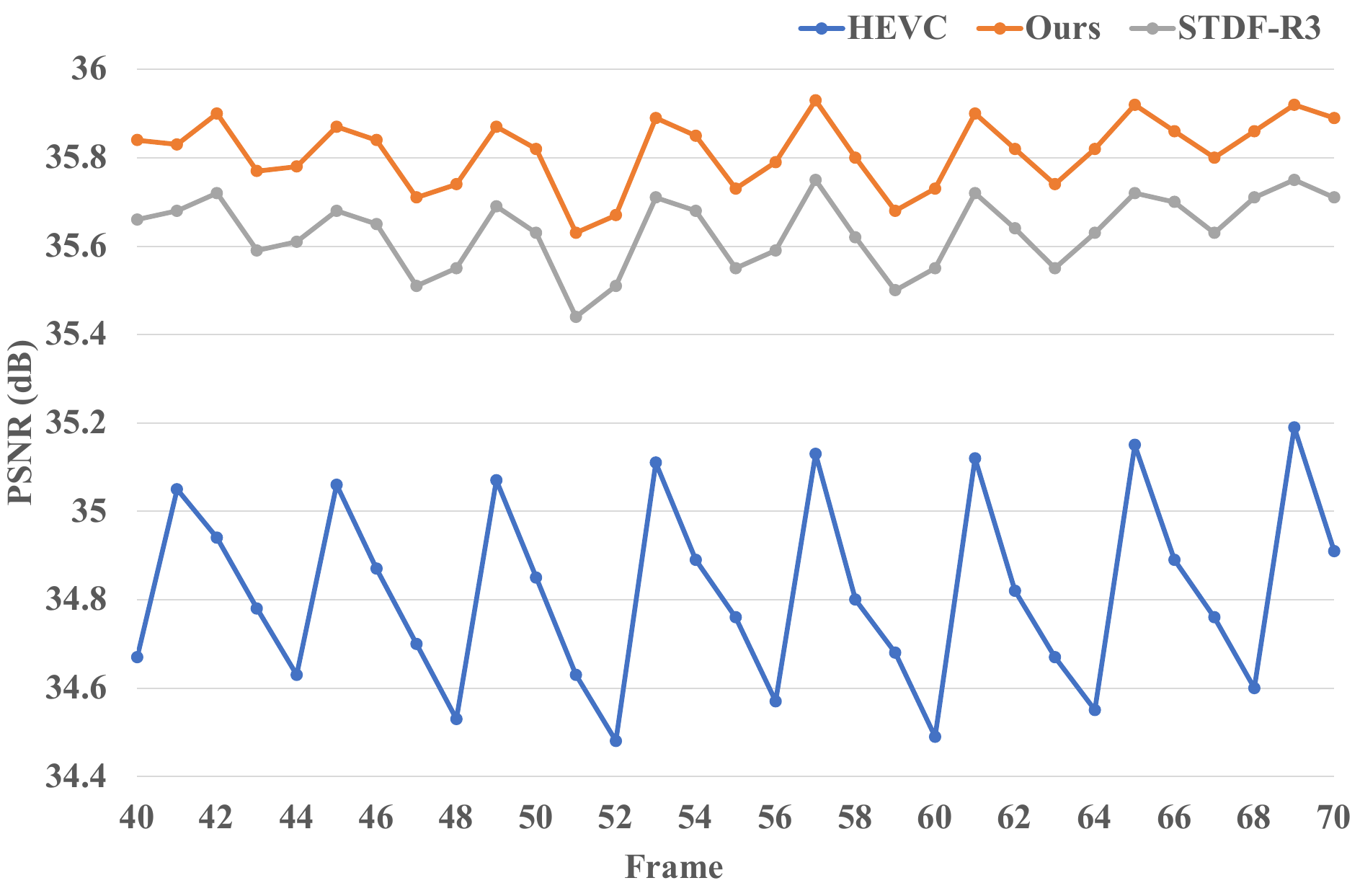}
\caption{\textbf{PSNR curves of HEVC baseline, STDF-R3 and our method
on FourPeople sequence at QP=37.} 
}
\label{fig:bodong}
\end{figure}
\textbf{Quality fluctuation.} 
To demonstrate that our method can improve the quality fluctuation, we plot PSNR curves of the FourPeople sequence in Figure \ref{fig:bodong}. As can be seen, our VIG can effectively enhance all the frames and reduce the gap between high-quality and low-quality frames. 

\begin{figure*}[htb]
\centering
\includegraphics[width=0.95\linewidth]{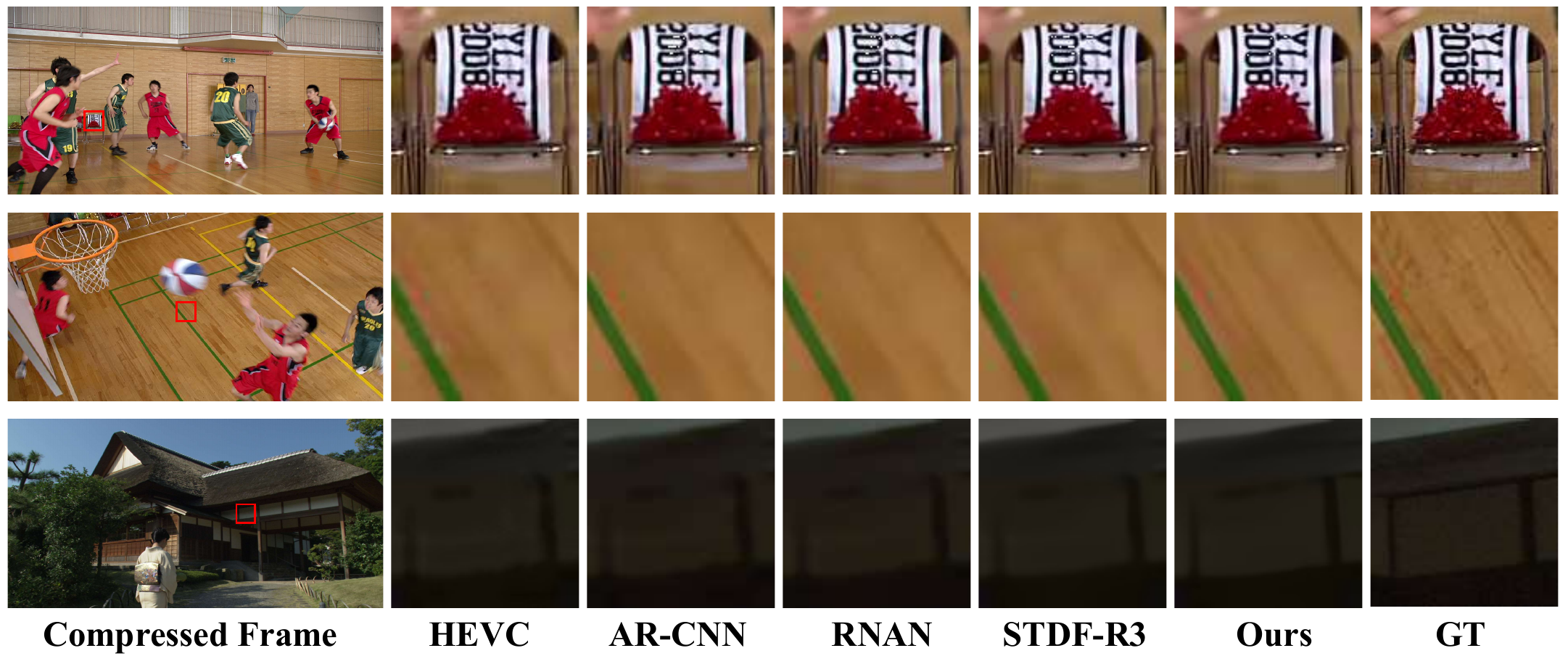} 
\caption{\textbf{Qualitative comparisons on the MFQE2.0.}}
\label{fig:zhuanguan}
\end{figure*}

\subsubsection{Qualitative Evaluation}
Qualitative comparisons are shown in Figure \ref{fig:zhuanguan}. Our VIG can recover finer details and sharper edges than AR-CNN \cite{arcnn}, RNAN \cite{rnan}, and STDF-R3 \cite{stdc}. 
For instance, only VIG successfully recover more apparent texture in the floor, more distinct letters on the chair, and more precise window borders than others.

\subsection{Ablation Study}
To better understand VIG, we ablate each critical component and evaluate the performance on the MFQE2.0 dataset.

\subsubsection{Components analysis of VIG}

In this part, various experiments illustrate the effectiveness of the proposed VIG. 
First, we validate the necessity of every component. Then, we investigate the impacts of different configurations of TGD.

\paragraph{Importance of components.}

\begin{table}[tb]
\centering
{
\resizebox{0.95\columnwidth}{!}
{%
\begin{tabular}{c| c c c c | c |c}
    \hline
    \multicolumn{1}{c|}{Method} &
    \multicolumn{1}{c}{Base} &
    \multicolumn{1}{c}{TF} &
    \multicolumn{1}{c}{UD} &
    \multicolumn{1}{c|}{TGD} &
    \multicolumn{1}{c|}{$\Delta$PSNR(dB)} &
    \multicolumn{1}{c}{$\Delta$SSIM($\times{10}^{-2}$)} \\
    \hline
    (i) & \checkmark  & &  &  & 0.62  & 1.18 \\
    (ii) & \checkmark & \checkmark  &  &  & 0.66 & 1.20 \\
    (iii) & \checkmark  & & \checkmark& & 0.76 & 1.45 \\
    CRF & \checkmark&\checkmark & \checkmark & &  0.77 & 1.48  \\
    \hline
    VIG &\checkmark &\checkmark &\checkmark & \checkmark  & 0.83  & 1.60 \\
    \hline
    \end{tabular}
    }
}
\caption{\textbf{Impact of components.} Ablation studies of each component are conducted to understand VIG better. $\checkmark$ means that VIG has the current module. }
\label{tab:ablation study of MBT}
\end{table}

In this section, we conduct several ablation studies as Table \ref{tab:ablation study of MBT} to validate the effectiveness of VIG and the necessity of every proposed module. 
We validate the necessity of the Temporal Fusion (TF) module, the Up-down sampling (UD) module, and the TGD strategy. TF denotes the use of the 3D-CNN layer. UD represents the module based on Pixel-shuffle and Pixel-unshuffle operation in Figure \ref{fig:RF}.
For better understanding, we start with the `Base', which denotes a simplified structure based on several ordinary convolution layers. 
Benefiting from the UD module, each block only needs one TF module to achieve accurate temporal alignment. It is easier to align on high-level semantic features that remove redundancy.
As we can see, the complete comparisons demonstrate that each proposed module has a significant performance improvement. 
Including all essential modules, VIG gets +33.87$\%$/+0.21 dB higher $\Delta$PSNR than Base. 
Notably, the performance gains +20.60$\%$/+0.13 dB with UD and +7.79$\%$/+0.06 dB with TGD, which suggests that valid information significantly influences VQE.

\begin{figure}[tb]
\centering
\includegraphics[width=1\linewidth]{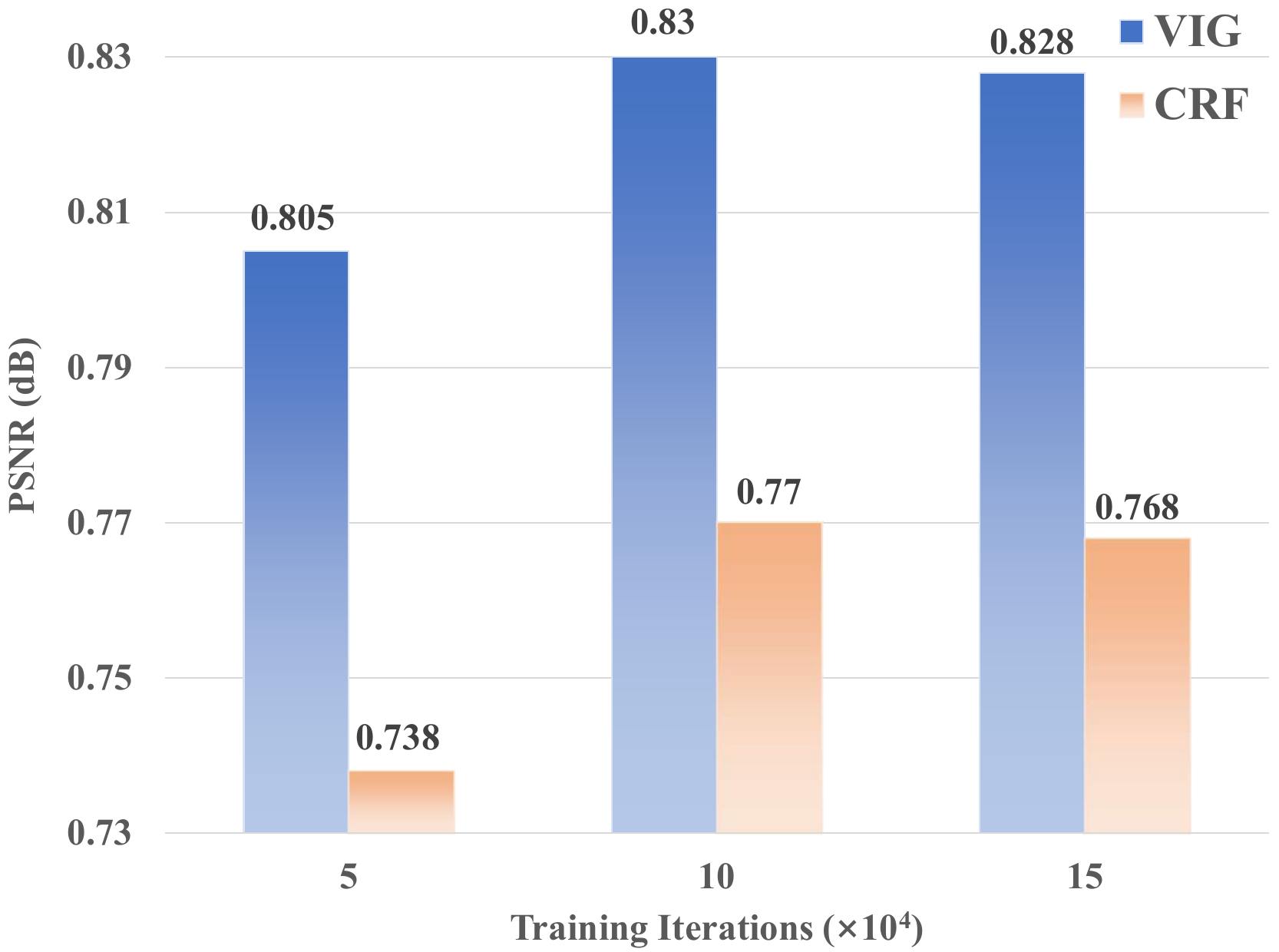}
\caption{\textbf{The impact of TGD.} 
}
\label{fig:shoulian}
\end{figure}

\paragraph{TGD accelerates convergence.} 
The performances of TGD in every stage are investigated in Figure \ref{fig:shoulian}. 
We train the contrast experiments for 120 thousand iterations, including 20 thousand iterations for pre-training and 100 thousand iterations for fine-tuning. It is worth noting that the model trained with TGD performs better than the model trained without it at every 50 thousand iterations. Although the pre-training stage takes up only 16.67$\%$ of the total training cost, the model still shows a stable advantage throughout the fine-tuning stage.


\subsubsection{Mask strategy analysis of TGD}
In this part, we exhibit extensive experiments to understand how the valid information of raw videos affects the final performance comprehensively. 
We first explore the impact of mask sizes on TGD. 
Then, we investigate the effect of mask ratios. Finally, we present how the multi-mask strategy affects the final performance.

\paragraph{Mask sizes.}
\begin{figure}[htb]
\centering
\includegraphics[width=1\linewidth]{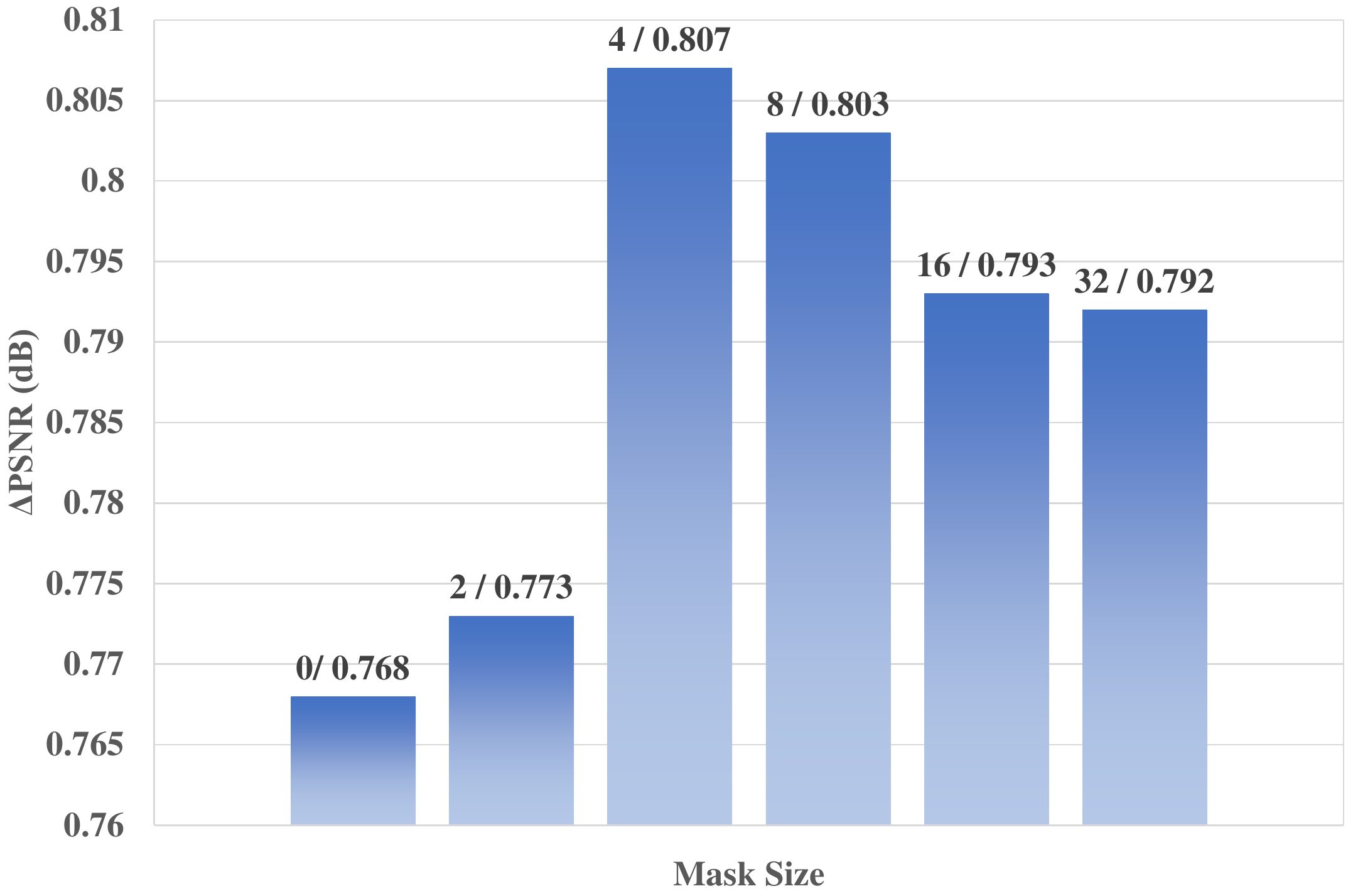}
\caption{\textbf{The impacts of different mask sizes.} The performances of different mask sizes are investigated on the premise that the ratio is selected as 0.9. For example, '4/0.807' denotes that $\mathit{w}$ is set to 4×4 in formula \ref{equ:mask} and PSNR increases by 0.807 dB.
For simplicity, we train the contrast experiments for 170 thousand iterations, including 20 thousand iterations for the pre-training stage and 150 thousand iterations for the fine-tuning stage. 
}
\label{fig:masksize}
\end{figure}

Here we try to understand which size can bring higher accuracy. 
As shown in Figure \ref{fig:masksize}, we run experiments based on changing $\mathit{w}$ in formula \ref{equ:mask}. 
Significantly, all the single-size mask strategies can increase performance. It may divert the network from paying attention to the recovery of details if utilizing too large a mask size, i.e., 16×16 and 32×32. 
Moreover, it also hurts the final performance when using an improper small size, which may lead the model to ignore the non-local features.
As a result, 4×4 is the most appropriate size, which can take care of both local features and non-local semantic information.

\paragraph{Mask ratios.}
\begin{figure}[htb]
\centering
\includegraphics[width=1\linewidth]{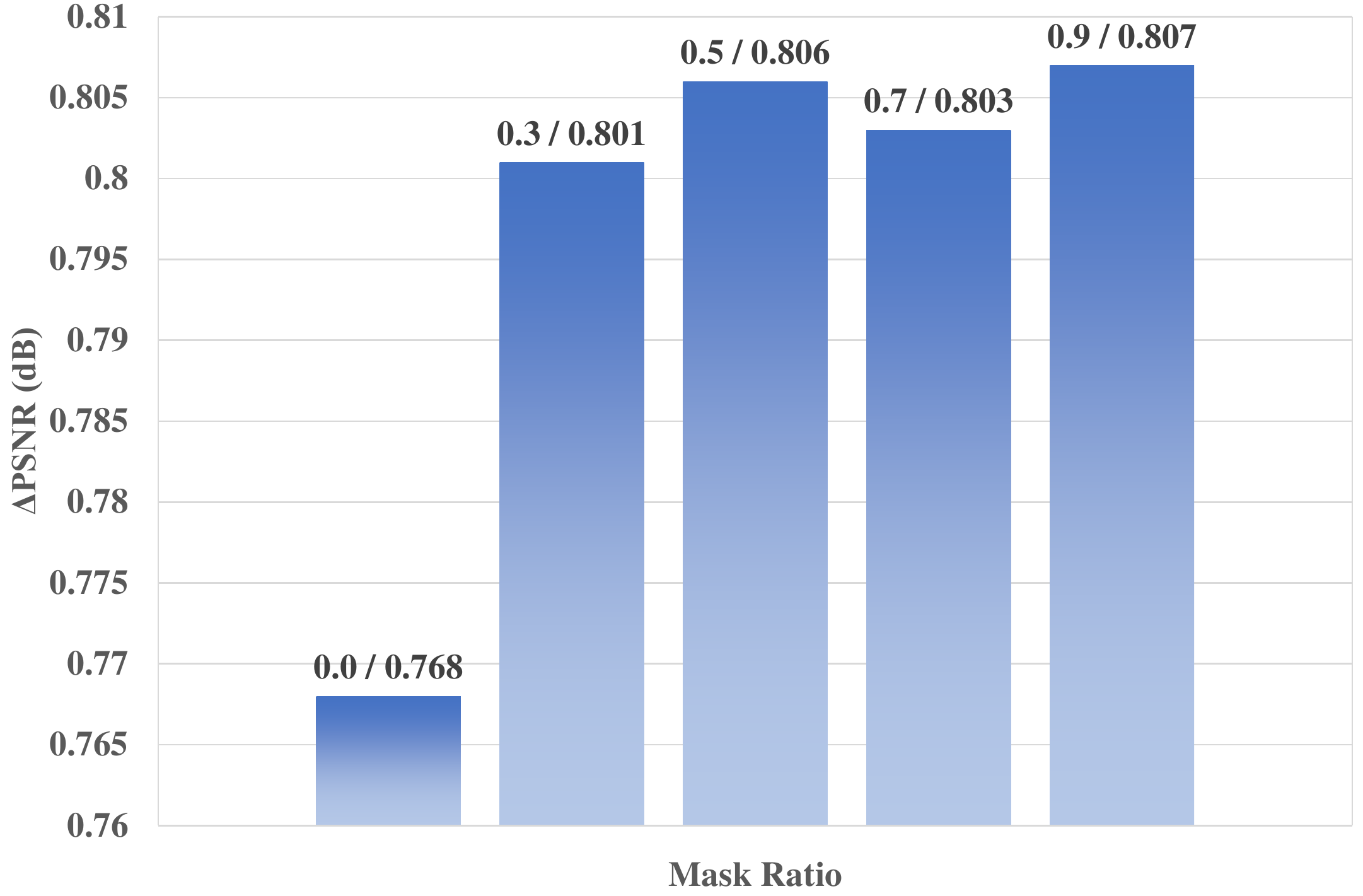}
\caption{\textbf{The impacts of different mask ratios.} The results of different mask ratios are studied based that the mask size is set to 4×4. For example, '0.9/0.807' means that $\mathit{r}$ is set to 0.9 in formula \ref{equ:mask} and PSNR increases by 0.807 dB. We also train the five contrast experiments for 170 thousand iterations as Figure \ref{fig:masksize}.
}
\label{fig:maskratio}
\end{figure}
We also study how different mask ratios affect the effectiveness. The fine-tuning accuracy of different approaches under multiple masking ratios is summarized in Figure \ref{fig:maskratio}. When the masked patch size of 4 is adopted, different ratios perform stably well on the abroad range of masking ratios, from 0.3 to 0.9. We hypothesize that the raw pixel relations may be valid enough. Thus it enforces the network to learn raw long-range connections, even when a low ratio is used (e.g., 0.3). Note that the most appropriate ratio is 0.9. In other words, the model performs best when we use the raw video almost exclusively as input, which is counter-intuitive.

\paragraph{Multi mask strategy.}

\begin{table}[htb]
\centering
{
\resizebox{0.95\columnwidth}{!}
{%
\begin{tabular}{c|  c c | c |c}
    \hline
    \multicolumn{1}{c|}{Method} &
    \multicolumn{1}{c}{Multi Mask} &
    \multicolumn{1}{c|}{Cosine Annealing} &
    \multicolumn{1}{c|}{$\Delta$PSNR(dB)} &
    \multicolumn{1}{c}{$\Delta$SSIM($\times{10}^{-2}$)} \\
    \hline
    (i)  &  &    & 0.81  & 1.54 \\
    (ii)  &  &\checkmark  & 0.82 & 1.58 \\
    \hline
    VIG &  \checkmark& \checkmark&  0.83  & 1.60 \\
    \hline
    
    \end{tabular}
    }
}
\caption{\textbf{Impact of Multi Mask.} Ablation studies of mask strategy are conducted to understand TGD better. $\checkmark$ means that VIG has the current setting. }
\label{tab:ablation study of multi}
\end{table}

As shown in Table \ref{tab:ablation study of multi}, we set the mask to 4 and ratio to 0.9 as the basic training strategy, which is the best combination among the single-mask strategies in Figure \ref{fig:masksize}. 
Then we use Cosine Annealing \cite{sgdr} to decay the learning rate from $3\times{10}^{-4}$ to ${10}^{-4}$ during both of the pre-training stage and the fine-tuning stage. 
Finally, we set the mask to 32, 16, 8, and 4 when using Cosine Annealing. Experimental results show that the multi-mask training strategy performs best.

\section{Conclusion}
In this paper, we propose a novel method to enhance compressed videos, whose main idea is to fully explore the valid clues from both compressed and raw videos. 
Specifically, CRF can capture the valid context of compressed videos by excluding redundant content.
Furthermore, TGD is proposed to help models better understand the raw pixel distribution based on raw videos. 
The extensive experiments demonstrate that our method can achieve superior performance over state-of-the-art methods. The proposed modules also can be easily adapted to existing multi-frame methods and video-related low-level tasks.

{\small
\bibliographystyle{ieee_fullname}
\bibliography{VSR}
}

\end{document}